\pdfoutput=1
\documentclass[11pt]{article}

\PassOptionsToPackage{table}{xcolor}
\usepackage[final]{acl}

\usepackage{times}
\usepackage{latexsym}
\usepackage[T1]{fontenc}
\usepackage[utf8]{inputenc}
\usepackage{microtype}
\usepackage{inconsolata}
\usepackage{graphicx}
\graphicspath{{figures/}}

\usepackage{amsmath,amssymb}
\usepackage{booktabs}
\usepackage{multirow}
\usepackage{array}
\usepackage{tabularx}
\usepackage{float}
\usepackage{enumitem}
\usepackage[breakable]{tcolorbox}
\usepackage{fontawesome5}
\usepackage{tikz}
\usetikzlibrary{positioning,shapes.geometric,shapes.callouts,arrows.meta}

\definecolor{lightergray}{HTML}{e0e0e0}
\definecolor{colLLM}{HTML}{2176AE}
\definecolor{coldLLM}{HTML}{E8651A}
\definecolor{colGold}{HTML}{4CAF50}
\definecolor{colTierDiff}{HTML}{F19065}   %
\definecolor{colTierAR}{HTML}{8569AC}     %
\definecolor{jcMercury}{HTML}{129A9A}
\definecolor{jcFlash}{HTML}{2A67E9}
\definecolor{jcMini}{HTML}{9CA1F2}
\definecolor{jcHaiku}{HTML}{D97757}
\definecolor{jcGPTbig}{HTML}{008EED}
\definecolor{jcGPTsmall}{HTML}{58BDEE}
\definecolor{jcDiffGemma}{HTML}{47a9ff}
\definecolor{jcNemotron}{HTML}{76B900}
\definecolor{jcGemma}{HTML}{93d2ff}

\providecommand{\phv}{\fontfamily{phv}\selectfont}
\definecolor{f2green}{HTML}{2E9E5B}
\definecolor{f2red}{HTML}{E23B2E}
\definecolor{f2pink}{HTML}{C0398C}

\usepackage{cleveref}

\title{Diffusion Large Language Models as Judges:\\ Accuracy, Latency and Biases}

\author{
  \textbf{Mirko Casu\textsuperscript{1,2}}\thanks{Work carried out while Mirko Casu was a visiting PhD student at EPFL.},
  \textbf{Sebastiano Battiato\textsuperscript{1}},
  \textbf{Andrea Cavallaro\textsuperscript{2}}
\\
  \textsuperscript{1}University of Catania, Italy \quad
  \textsuperscript{2}EPFL, Switzerland
\\
  \small{\texttt{mirko.casu@phd.unict.it}, \texttt{sebastiano.battiato@unict.it}, \texttt{andrea.cavallaro@epfl.ch}}
}

\begin{document}
\maketitle

\begin{abstract}
LLM judges, Large Language Models that evaluate the outputs of other language models, suffer from evaluation biases, such as position and verbosity. Autoregressive LLMs, today's standard, read their input sequentially, whereas Diffusion Large Language Models (dLLMs) can read it with bidirectional attention. Hence, dLLMs are strong candidates as judges. In this work we present the first evaluation of dLLMs as judges on nine evaluation biases. Furthermore, we introduce paradigm bias, a judge's preference for or against responses generated by models of its own paradigm (diffusion or autoregressive, in our case). We test three diffusion judges, Mercury-2, LLaDA-2.1-flash, and DiffusionGemma, against three capability-matched autoregressive ones, Claude-Haiku-4.5, Nemotron-3-Super, and Gemma-4-26B-A4B. On human agreement the two paradigms are statistically equivalent, while in every pair the dLLM is at least as precise and as robust as the autoregressive judge, and in the pair that shares an architecture it runs five to seven times faster. Two of the six judges also give lower scores to responses from their own paradigm, and self-preference bias appears in both paradigms. We release code, verdicts and anonymised annotations at \url{https://github.com/epfl-mints/dllms-as-judges}.
\end{abstract}

\section{Introduction}
\label{sec:introduction}

An LLM-as-a-judge is a language model employed to evaluate data produced by other language models~\cite{zheng2023judging,li2025generation,bavaresco2025llmsinsteadhumanjudges}. An LLM-as-a-judge is defined by a task to follow, responses to evaluate, and a grading scheme called a \emph{rubric}. LLM judges can work \emph{pairwise}, giving a preference (or a tie) between two candidate responses~\cite{zheng2023judging,wang2023notfairevaluators,liusie2024comparative}, or \emph{pointwise}, giving a numeric score to each response~\cite{ye2024justiceprejudicequantifyingbiases,gu2025surveyllmasajudge}.

These judges are vulnerable to \emph{evaluation biases}~\cite{ye2024justiceprejudicequantifyingbiases}, which follow the pattern of human cognitive biases. They are systematic deviations that lead a judge to preferences shaped by information beyond the content of a response. Biases include \emph{position} bias, preferring a response by where it appears~\cite{wang2023notfairevaluators,shi2025judgingjudgessystematicstudy}, \emph{verbosity} bias, favouring a response by its length~\cite{saito2023verbositybiaspreferencelabeling,hu2024length}, and \emph{self-preference}, preferring responses generated by the same model as the judge~\cite{wataoka2024self,panickssery2024llmevaluatorsrecognizefavor}. Other biases include favouring responses that carry a bibliographic source~\cite{chen-etal-2024-humans}, or that report content many people hold to be true~\cite{ma2025judgingmindsperspectivesmean}. The emotional tone of a response influences the verdict~\cite{feuer2025style}. CALM also counts as a bias the failure to spot logical fallacies or faulty reasoning in a response, even where they are visible~\cite{ye2024justiceprejudicequantifyingbiases}.

Diffusion Large Language Models (dLLMs)~\cite{Yi2024Diffusion,li2026surveydiffusionlanguagemodels} are a novel class of language models that differ from their autoregressive counterpart (today's standard, the LLM) in the attention mechanism. A dLLM can read its input with \emph{bidirectional attention}~\cite{nie2025llada,li2026surveydiffusionlanguagemodels}, where every token attends to the tokens before and after it. An LLM judge also sees all the responses when it writes its verdict, but the first response cannot attend to the second. This feature may reduce the exposure to evaluation biases for dLLMs used as judges, above all for the biases that turn on the position and the length of a response, though this remains an untested hypothesis. In the two open diffusion judges of our panel, attention is not fully bidirectional. DiffusionGemma encodes the prompt with a causal encoder and denoises a block of $256$ tokens~\cite{diffusiongemmateam2026diffusiongemmatechnicalreport}, and LLaDA-2.1-flash uses block-wise causal attention~\cite{bie2026llada21speedingtextdiffusion}. Lower exposure to bias matters because judges also annotate the data that trains new models~\cite{zheng2023judging,bavaresco2025llmsinsteadhumanjudges,gu2025surveyllmasajudge}, so a judge's biases may impact other models trained on those data~\cite{singhal2023longway,park2024disentangling}. The most recent dLLMs, among them Mercury~\cite{labs2025mercuryultrafastlanguagemodels}, LLaDA~\cite{bie2026llada21speedingtextdiffusion}, and DiffusionGemma~\cite{diffusiongemmateam2026diffusiongemmatechnicalreport}, now reach sizes and benchmark scores close to their autoregressive counterparts. So far they have been tested only as generators, on their perplexity~\cite{lou2024sedd,sahoo2024mdlm} and their adversarial robustness~\cite{zhang2025jailbreakinglargelanguagediffusion,Wen2025The}.

We therefore present the first evaluation of dLLMs as judges, comparing three diffusion judges (Mercury-2, LLaDA-2.1-flash, DiffusionGemma) against three autoregressive ones (Claude-Haiku-4.5, Nemotron-3-Super, Gemma-4-26B-A4B) on nine evaluation biases~\cite{ye2024justiceprejudicequantifyingbiases}. We also measure the Accuracy, Precision, and Robustness of each judge, which then combine into a judge profile. Furthermore, we identify a new bias, \emph{paradigm bias}, the tendency to prefer (or to mark down) responses generated by models of a judge's own paradigm (the way a model generates text), diffusion and autoregressive in the case of our panel. We quantify paradigm bias with a regression developed for self-preference and \emph{family bias}~\cite{spiliopoulou2025playfavoritesstatisticalmethod}, the preference for responses generated by a judge's own model family~\cite{li2026preferenceleakagecontaminationproblem,koyama2026mllmasajudgeexhibitsmodelpreference} (Claude Opus preferring an answer from Claude Sonnet over one from GPT, for instance). With the same regression we measure a judge's self-preference bias.

\noindent In this paper we show that:
\begin{itemize}
\item each dLLM judge is at least as precise and as robust as the capability-matched autoregressive one it is paired with, and on Accuracy the two paradigms are equivalent within a margin of $.05$ (gap $-.003$, $95\%$ CI $[-.042,+.045]$);
\item a regression against a reference built from the other judges can estimate paradigm bias with no human labels, on a panel where every model both answers and judges;
\item paradigm bias is a pattern of single judges in our panel, with two of the six marking down responses from their own paradigm, against what the literature on family bias suggests.
\end{itemize}
Moreover, we present the first comparison of two judges that share an architecture and differ in paradigm, DiffusionGemma and Gemma-4-26B-A4B, where their paradigm biases differ.
\section{Metrics}
\label{sec:evaluation}

We evaluate the performance of each judge with three metrics, namely Accuracy, Precision, and Robustness. A judge can return answer A, answer B, or a tie, where the two answers are of equal quality. Precision and Robustness take that verdict from a pairwise query, and Accuracy reads it off two pointwise scores. Each metric compares a pair of verdicts $y$ and $y'$. Precision and Robustness count $1$ when the two verdicts match and $0$ otherwise. Accuracy grades the pair with a weight $w(y,y')$, where a match scores $1$, a tie against a decisive call $\tfrac{1}{2}$, and a reversal $0$.

\textit{Accuracy} ($\mathrm{A}$) measures the agreement between a judge and humans. Let $\mathcal{H}$ be the human annotations, and for $a \in \mathcal{H}$ let $y^a$ and $h^a$ be the judge's and the annotator's verdict. Write $\mathrm{A}_o = \frac{1}{|\mathcal{H}|}\sum_{a} w(y^a,h^a)$ for the observed agreement, and $\mathrm{A}_e = \sum_{y,y'} p(y)\,q(y')\,w(y,y')$ for the same quantity under chance, where $p$ and $q$ are the judge's and the annotators' verdict frequencies. Accuracy is a linearly weighted $\kappa$~\cite{cohen1968weighted}:
\begin{equation}
\mathrm{A} = \frac{\mathrm{A}_o - \mathrm{A}_e}{1 - \mathrm{A}_e}.
\label{eq:accuracy}
\end{equation}

The score is calibrated so that chance gives $0$ and perfect agreement $1$. The correction matters here because the two sides need not use the tie at the same rate, and without it a judge would gain from tying more often than the annotators it is measured against. Every judge is scored on the same annotations, one from each annotator on every pair, and we report a bootstrap standard error clustered on the source prompts.

\textit{Precision} ($\mathrm{P}_b$) measures the judge's consistency with itself by scoring the same item twice, without any perturbation. For each item $i \in I_b$ (the items for bias $b$ in their original form, before the bias-specific perturbation defined under Robustness below), we draw a second independent verdict $y_i^{(2)}$ under identical conditions. $\mathrm{P}_b$ is the mean agreement between the two:
\begin{equation}
\mathrm{P}_b = \frac{1}{|I_b|}\sum_{i \in I_b} \mathbf{1}\!\left[y_i^{(1)} = y_i^{(2)}\right],
\label{eq:precision}
\end{equation}
where $y_i^{(1)},y_i^{(2)} \in \{A, B, \text{tie}\}$ are the first and second verdicts.

\definecolor{annBlue}{HTML}{2A9D8F}
\definecolor{annOchre}{HTML}{E76F51}
\definecolor{verdictBox}{HTML}{F4F2EF}
\newcommand{\pillA}[1]{\tcbox[on line,colback=annBlue!12,colframe=annBlue,boxrule=.7pt,arc=2.5pt,left=3pt,right=3pt,top=1pt,bottom=1pt,boxsep=0pt]{\phv\scriptsize\bfseries\color{annBlue!85!black}#1}}
\newcommand{\pillB}[1]{\tcbox[on line,colback=annOchre!14,colframe=annOchre,boxrule=.7pt,arc=2.5pt,left=3pt,right=3pt,top=1pt,bottom=1pt,boxsep=0pt]{\phv\scriptsize\bfseries\color{annOchre!85!black}#1}}
\newcommand{\pillchg}[1]{\tcbox[on line,colback=red!12,colframe=red!70!black,boxrule=.7pt,arc=2.5pt,left=3pt,right=3pt,top=1pt,bottom=1pt,boxsep=0pt]{\phv\scriptsize\bfseries\color{red!75!black}#1}}
\newcommand{\dangertri}{\tikz[baseline=-0.3ex,line join=round]{%
  \fill[red!80!black,rounded corners=1.1pt] (-1.0em,0) -- (1.0em,0) -- (0,1.65em) -- cycle;
  \fill[white,rounded corners=0.3pt] (-0.105em,0.62em) rectangle (0.105em,1.15em);
  \fill[white] (0,0.40em) circle (0.115em);}}
\newlength{\fcm}\setlength{\fcm}{5pt}
\newlength{\fccap}\newlength{\fcstrutdp}\newlength{\fcpilldp}
\newsavebox{\fcbox}
\newcommand{\fctext}[2]{\begin{minipage}[t]{#1}#2\par\kern-\prevdepth\end{minipage}}
\newcommand{\fcverdict}[1]{\settoheight{\fccap}{\bfseries H}\setlength{\fcstrutdp}{\dp\strutbox}%
  \settodepth{\fcpilldp}{\pillA{1}}\sbox{\fcbox}{#1}%
  \raisebox{0pt}[\fccap][\dimexpr\dp\fcbox-\fcstrutdp+\fcpilldp\relax]{\usebox{\fcbox}}}
\begin{figure}[t]
\centering
\includegraphics[width=\linewidth]{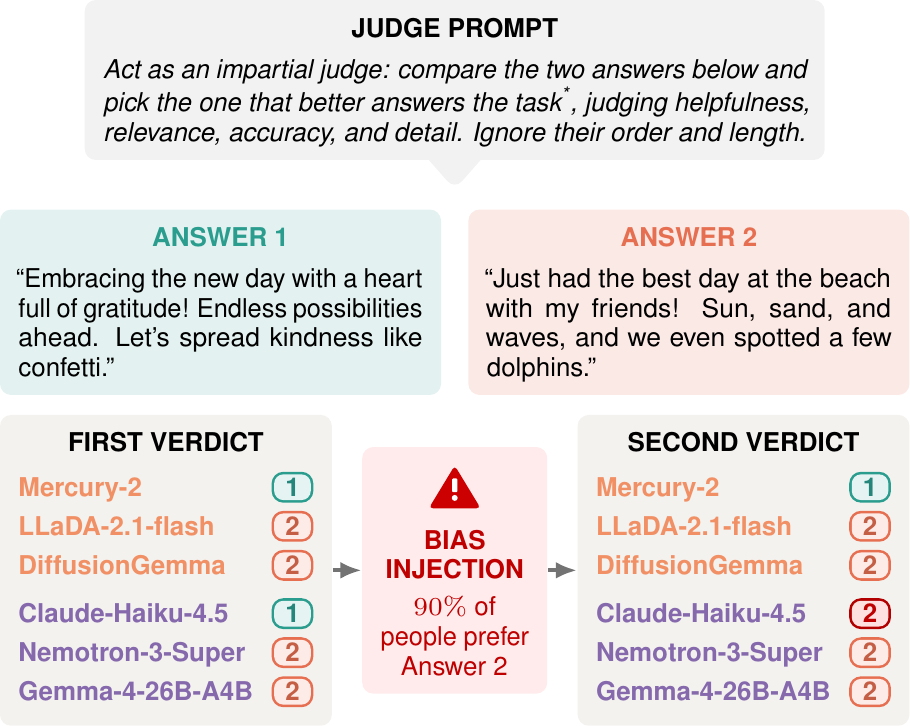}
\caption{A bandwagon comparison, and how Robustness counts it. The top box is the judge's pairwise prompt (condensed); the judge compares the two answers and returns a verdict. Six judges give a first verdict $y_k^{(1)}$ (the pill is each judge's pick). We then inject the bandwagon cue, a fabricated majority claiming that $90\%$ of people prefer Answer 2, and read the verdict after it, $\hat{y}_k^{(b)}$, where a \textcolor{red!75!black}{red} pill marks one that changed. Each judge contributes one comparison to Eq.~\ref{eq:stability}: five judges hold and count $1$, while Claude-Haiku-4.5 flips and counts $0$. \textsuperscript{*}The task is to write a positive tweet.}
\label{fig:bandwagon}
\end{figure}

\textit{Robustness} ($\mathrm{R}_b$) measures whether a judge holds its verdict when the input changes in a way that does not change which answer is better, such as swapping the order of the two answers. Here each perturbation is bias-specific, and goes either into a candidate answer or into the judge's instruction. Let $D_b$ be the set of perturbed comparisons for bias $b$, obtained by perturbing the items of $I_b$. A single item may generate multiple perturbed comparisons, one for each perturbation variant considered. For each comparison $k \in D_b$, let $y_k^{(1)} \in \{A, B, \text{tie}\}$ be the verdict before perturbation and $\hat{y}_k^{(b)}$ the verdict after. $\mathrm{R}_b$ is the mean agreement between the two:
\begin{equation}
\mathrm{R}_b = \frac{1}{|D_b|}\sum_{k \in D_b} \mathbf{1}\!\left[y_k^{(1)} = \hat{y}_k^{(b)}\right].
\label{eq:stability}
\end{equation}
Figure~\ref{fig:bandwagon} works one such comparison through for all six judges. Where a bias carries a single variant, $D_b = I_b$; otherwise $|D_b| > |I_b|$. The biases also split on what the right verdict is after the perturbation. For most of them the perturbation leaves the better answer unchanged, so holding the verdict is correct. For one it makes the better answer worse, so there a changed verdict is correct and we report the detection rate $1 - \mathrm{R}_b$ (higher is better). \Cref{sec:biases} names them. Precision and Robustness follow CALM's Consistency Rate and Robustness Rate~\cite{ye2024justiceprejudicequantifyingbiases}, respectively.

We combine the three measures into a single judge-level index for ranking and model selection. In place of Accuracy we take $\mathrm{A}^{+} = \max(0,\, \mathrm{A})$, which gives $0$ to a judge that scores below chance. Let $\bar{\mathrm{P}} = \frac{1}{|B|}\sum_{b \in B} \mathrm{P}_b$ be the mean Precision over all biases $B$, and let $\bar{\mathrm{R}} = \frac{1}{|B|}\sum_{b \in B} \mathrm{R}_b$ be the mean Robustness over the same biases, where we count the inverted bias as its detection rate $1-\mathrm{R}_b$, so that higher is better throughout. The \textit{judge profile score} is their geometric mean:
\begin{equation}
\mathrm{J} = \bigl(\mathrm{A}^{+} \cdot \bar{\mathrm{P}} \cdot \bar{\mathrm{R}}\bigr)^{1/3}.
\label{eq:judge_score}
\end{equation}
An ideal judge scores $1$, being perfectly human-aligned, deterministic, and immune to perturbations. The geometric mean pulls $\mathrm{J}$ toward zero if a factor is weak.

\section{Paradigm-bias regression}
\label{sec:paradigm}

We introduce \emph{paradigm bias} as a per-judge coefficient that shows whether a judge systematically scores answers from its own paradigm higher or lower. The family-bias regression of~\citet{spiliopoulou2025playfavoritesstatisticalmethod} scores a judge's preference for a whole group of models (a family) through a dedicated coefficient on a same-group indicator. Fitted jointly over all items $i$, judges $j$, quality dimensions $d$, and evaluated models $m$, the model reads

\begin{equation}
\begin{aligned}
\tilde S_{idmj} \;=\;& \alpha \;+\; \underbrace{\delta_j + \beta_j\, S_{idm}}_{\text{Human alignment}} \;+\; \underbrace{\gamma_j\, \mathbf{1}_j(m)}_{\text{Self-bias}} \\[2pt]
&+\; \underbrace{\lambda_{F(j)}\, \mathbf{1}_{F(j)}(F(m))}_{\text{Family bias}} \;+\; \eta_d \;+\; \epsilon_{idmj},
\end{aligned}
\label{eq:playfav}
\end{equation}
with $S_{idm}$ an independent reference score, human annotations in their main analysis, and $F(\cdot)$ the model family. They set $\mathbf{1}_{F(j)}(F(j))=0$. The family term then leaves out the judge's own completions and leaves them to $\gamma_j$. We make three changes:
\begin{itemize}
    \item We replace the family indicator with the paradigm indicator $P$, turning the group coefficient into $\lambda_P$.
    \item We replace the human reference $S_{idm}$ with the leave-one-judge-out consensus $S_{-j}(i,m)$ of Eq.~\ref{eq:consensus}, which needs no item-level human labels. They run a similar substitution as a robustness check, leaving out one family at a time; we make the cross-judge reference primary and leave out one judge.
    \item We fit one regression per judge, where they fit a single joint model; the per-judge coefficients carry no $j$ subscript.\footnote{With $j$ fixed within a fit the intercepts $\alpha$ and $\delta_j$ merge, and the per-dimension term $\eta_d$ drops, since our ratings carry no quality-dimension index.}
\end{itemize}

Specifically, for each judge we fit an ordinary least squares regression over its pointwise ratings $\tilde S_{imj}$ of model $m$'s answer to prompt $i$:

\begin{equation}
\begin{aligned}
\tilde S_{imj} \;=\;& \underbrace{\alpha + \beta\, S_{-j}(i,m)}_{\substack{\text{Cross-judge}\\\text{alignment}}} \,+\, \underbrace{\gamma_{\text{self}}\, \mathbf{1}[m = j]}_{\text{Self-bias}} \\[2pt]
&+\, \underbrace{\lambda_P\, \mathbf{1}[P(m) = P(j),\, m \neq j]}_{\substack{\text{Paradigm}\\\text{bias}}} \,+\, \epsilon_{imj}.
\end{aligned}
\label{eq:archbias}
\end{equation}
The first block ties each rating to the \emph{cross-judge consensus}

\begin{equation}
S_{-j}(i,m) \;=\; \frac{1}{|\mathcal{J}|-1}\sum_{k\in\mathcal{J},\,k\neq j} S(i,m,k),
\label{eq:consensus}
\end{equation}
the mean score the other judges in the panel $\mathcal{J}$ give the same answer; $\beta$ measures how strongly the judge's ratings move with that consensus, and using it as the reference lets the regression run without item-level human labels. The self-bias term $\gamma_{\text{self}}\,\mathbf{1}[m=j]$ is the extra credit a judge gives its own outputs. The paradigm-bias term $\lambda_P\,\mathbf{1}[P(m)=P(j),\,m\neq j]$ is the extra credit it gives same-paradigm answers that are not its own, with $P\in\{\text{diffusion},\text{autoregressive}\}$. The $m\neq j$ clause carries over the exclusion of Eq.~\ref{eq:playfav}. The two indicators never fire together, and the regression estimates self-preference and paradigm preference separately. A positive $\lambda_P$ means the judge favours its own paradigm, a negative one that it marks that paradigm down, and $\epsilon_{imj}$ is the residual.

\section{Biases and datasets}
\label{sec:biases}

We assess our judges on the nine perturbation-based biases of CALM~\cite{ye2024justiceprejudicequantifyingbiases}, using the datasets and prompts the authors release\footnote{CALM's official repo:~\url{https://github.com/Y0oMu/LLM-Judge-Bias-Dataset}.}. On all nine the judges work in a pairwise setting (Figure~\ref{fig:prompts} in Appendix~\ref{sec:appendix:templates}) and return a verdict that gives the preference to answer [A] or [B]. The CALM prompt offers no tie here, and the judges return one in $7$ of the $75{,}766$ verdicts we collect. We present each judge both with unperturbed answer pairs, which give us Precision, and with pairs that carry a perturbed answer, which give us Robustness. The perturbation goes either into the answers (authority, sentiment, verbosity, fallacy-oversight) or into the prompt (position, distraction, compassion-fade, bandwagon, diversity). Position, verbosity, and fallacy-oversight carry one perturbation variant per item, and the rest carry several. Fallacy-oversight is also the one bias whose perturbation makes the better answer worse, so a judge that changes its verdict there is right and we report its detection rate. Sample sizes vary from $150$ to $500$ items per bias, following the CALM authors (Appendix~\ref{sec:appendix:datasets}).\footnote{CALM's Table 2 lists $500$ items for sentiment, where the dataset they release holds $150$. We follow the released data.}

We build the paradigm panel by cross-scoring the six judges against their own answers and against one another's. Each judge first answers each of the $150$ prompts of Table~\ref{tab:datasets} in Appendix~\ref{sec:appendix:datasets} (the \textit{ours} row), and then rates every answer in the panel under the pointwise $1$--$10$ format (Figure~\ref{fig:prompts} in Appendix~\ref{sec:appendix:templates}, right). This gives us $5400$ pointwise scores (one has no readable score), $900$ per judge, $150$ of them on the judge's own answers. Every model is both a generator and a judge. From this matrix we read both the self-bias and the paradigm bias (Eq.~\ref{eq:archbias}).

\section{Results}
\label{sec:results}

\subsection{Panel}
\label{sec:methods}

\begin{table*}[t]
\centering
\small
\setlength{\tabcolsep}{4pt}
\arrayrulecolor{black}
\begin{tabular}{lllrl}
\toprule
Judge & Lineage & Type & Params (B) & Source \\
\midrule
\textcolor{colTierDiff}{Mercury-2}~\cite{inceptionlabs2026mercury2}          & Mercury  & n/d$^\dagger$ & n/d$^\dagger$ & Closed \\
\textcolor{colTierDiff}{LLaDA-2.1-flash}~\cite{bie2026llada21speedingtextdiffusion}    & LLaDA    & MoE & 100 (6.1) & Open \\
\textcolor{colTierDiff}{DiffusionGemma}~\cite{diffusiongemmateam2026diffusiongemmatechnicalreport}     & Gemma    & MoE & 25.2 (3.8) & Open \\
\midrule
\textcolor{colTierAR}{Claude-Haiku-4.5}~\cite{anthropic2025haiku45}   & Claude   & n/d$^\dagger$ & n/d$^\dagger$ & Closed \\
\textcolor{colTierAR}{Nemotron-3-Super}~\cite{nvidia2026nemotron3super}   & Nemotron & MoE & 120 (12)  & Open \\
\textcolor{colTierAR}{Gemma-4-26B-A4B}~\cite{google2026gemma4}    & Gemma    & MoE & 25.2 (3.8) & Open \\
\bottomrule
\end{tabular}
\caption{The six-judge panel, coloured by paradigm (\textcolor{colTierDiff}{diffusion} top, \textcolor{colTierAR}{autoregressive} bottom). Params are total (active) weights in billions; n/d marks counts not disclosed by the vendor ($^{\dagger}$).}
\label{tab:models_full}
\end{table*}

We chose the three strongest publicly available Diffusion Large Language Models and paired each with an autoregressive one of matching capability (full panel in Table~\ref{tab:models_full}). We paired Mercury-2 with Claude-Haiku-4.5 following the benchmarks Mercury's developers publish~\cite{inceptionlabs2026mercury2}. The other two pairs go by parameter count, MoE design, and release window. LLaDA-2.1-flash goes with Nemotron-3-Super ($\sim$100--120B, released a month apart in early 2026), and DiffusionGemma with Gemma-4-26B-A4B, both built on the same $26$B-A$4$B Gemma 4 MoE architecture. \Cref{sec:appendix:capability} documents the match. Every judgment we evaluate runs at temperature $0.7$, the setting of CALM~\cite{ye2024justiceprejudicequantifyingbiases}, with the other sampling settings left at each model's defaults (\Cref{sec:appendix:decoding}). We chose $0.7$ over greedy decoding because DiffusionGemma degenerates at temperature $0$ ($80\%$ of its answers cannot be parsed) and the Mercury-2 API accepts only $0.5$ or above.

\subsection{Human reference}
\label{sec:methods:human}

Six human annotators and all six judges scored the same $120$ answer pairs, blind to the identity of the generating models, to their paradigm, and to the source dataset. We drew twenty prompts from the $150$ of Table~\ref{tab:datasets} in Appendix~\ref{sec:appendix:datasets} (the \textit{ours} row), all of them from the Emerton pool, and took as answers the six the panel models wrote for them, the same generations the judges rate pointwise in \Cref{sec:results:regression}. For each prompt we arranged the six answers in a circle and paired every answer with its two neighbours, which gives six pairs per prompt. We also shuffled the order of the responses for each annotator.

One annotator agreed with the other five less than they agreed among themselves ($.49$ against $.56$), so we left that annotator out. The five who remain agree moderately (Krippendorff $\alpha\!=\!.47$, with each verdict coded as the model whose answer it prefers, or a tie~\cite{krippendorff2018content}). Most of what they disagree on is the tie: $63\%$ of the pairs they read differently pit a tie against a decisive call. Where both pick a side they agree well (Cohen $\kappa$~\cite{cohen1960coefficient} from $.59$ to $.83$, mean $.68$). Those five are our \emph{soft} human reference, and we score every judge against each of their $120 \times 5 = 600$ annotations to obtain Accuracy (Eq.~\ref{eq:accuracy}).

\subsection{Accuracy}
\label{sec:results:accuracy}

\begin{table*}[t]
\centering
\small
\renewcommand{\arraystretch}{1.15}
\begin{tabular}{llcccc}
\toprule
Judge & Par. & $\mathrm{A}$ & $\bar{\mathrm{P}}$ & $\bar{\mathrm{R}}$ & \textit{J} \\
\midrule
\textcolor{colTierDiff}{DiffusionGemma} & \textcolor{colTierDiff}{\textsc{di}} & $.32\ (.04)$ & $.969\ (.004)$ & $.879\ (.005)$ & \textit{.65}\ (.03) \\
\textcolor{colTierAR}{Gemma-4-26B-A4B} & \textcolor{colTierAR}{\textsc{ar}} & $.30\ (.06)$ & $.955\ (.006)$ & $.877\ (.006)$ & \textit{.63}\ (.04) \\
\textcolor{colTierDiff}{Mercury-2} & \textcolor{colTierDiff}{\textsc{di}} & $.27\ (.06)$ & $.931\ (.006)$ & $.868\ (.006)$ & \textit{.60}\ (.04) \\
\textcolor{colTierAR}{Claude-Haiku-4.5} & \textcolor{colTierAR}{\textsc{ar}} & $.30\ (.05)$ & $.807\ (.009)$ & $.742\ (.008)$ & \textit{.57}\ (.03) \\
\textcolor{colTierDiff}{LLaDA-2.1-flash} & \textcolor{colTierDiff}{\textsc{di}} & $.22\ (.05)$ & $.910\ (.008)$ & $.852\ (.007)$ & \textit{.56}\ (.04) \\
\textcolor{colTierAR}{Nemotron-3-Super} & \textcolor{colTierAR}{\textsc{ar}} & $.21\ (.05)$ & $.901\ (.008)$ & $.840\ (.007)$ & \textit{.54}\ (.05) \\
\bottomrule
\end{tabular}
\caption{Judge profile scores, ranked by decreasing J (Eq.~\ref{eq:judge_score}); higher is better. \textit{Par.} gives the paradigm, \textcolor{colTierDiff}{\textsc{di}} for diffusion and \textcolor{colTierAR}{\textsc{ar}} for autoregressive. $\mathrm{A}$ is chance-corrected (Eq.~\ref{eq:accuracy}). It runs on a lower scale than the two rates beside it, with $0$ as the floor where they have $.5$. Parenthesised values are bootstrap standard errors ($10{,}000$ resamples); every figure is written to the precision its standard error supports.}
\label{tab:judge_score}
\end{table*}

On Accuracy the two paradigms are equivalent within a margin of $.05$ (gap $-.003$, $95\%$ CI $[-.042,+.045]$; \Cref{sec:robustness}). The ranking puts DiffusionGemma first and Gemma-4-26B-A4B second, one diffusion judge and one autoregressive judge (Table~\ref{tab:judge_score}); the two are the smallest models whose size the vendor discloses, and the only two in the panel that share an architecture. Accuracy is low to moderate across the panel, from $.212$ to $.316$, and the annotators do not agree much better among themselves. We score each annotator in the same way as a judge, against the other four, and the five reach $.299$ to $.407$ (mean $.341$). The best judge, DiffusionGemma ($.316$), is inside this range, so it agrees with the annotators about as much as they agree with each other. Their disagreement sets the ceiling on this task. Much of what holds the numbers down is the tie. The annotators tie on $.23$ of the items, the judges on $.22$ to $.42$, and five of the six judges tie more often than the annotators, the diffusion judges most of all. On the pairs where the judge and the annotator both pick a side, agreement climbs to $.69$--$.75$ (each judge on its own decisive subset, $278$ to $382$ items), and on these pairs DiffusionGemma stays first under three references (the five annotators, all six, and their majority label) while the middle of the ranking moves.

\subsection{Precision}

\begin{figure}[t]
\centering
\includegraphics[width=\columnwidth]{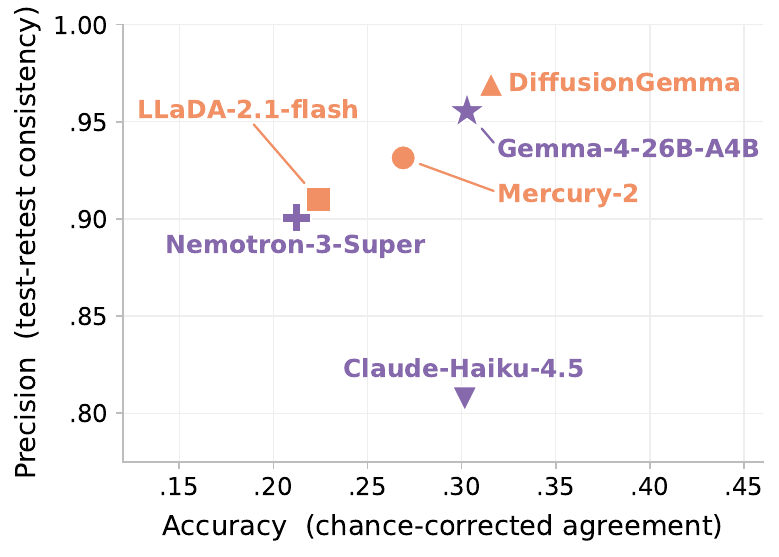}
\caption{Precision against Accuracy per judge (higher is better), coloured by paradigm (\textcolor{colTierDiff}{diffusion}, \textcolor{colTierAR}{autoregressive}). Each diffusion judge is at least as precise as the autoregressive judge it is paired with; the paradigms overlap on accuracy. Values in Table~\ref{tab:judge_score}.}
\label{fig:acc_prec}
\end{figure}

Each diffusion judge is at least as precise as its autoregressive counterpart. There is no clean split by paradigm, since Gemma-4-26B-A4B ranks below DiffusionGemma and above Mercury-2 and LLaDA-2.1-flash (Figure~\ref{fig:acc_prec}, Table~\ref{tab:judge_score}). Precision is close to perfect where the answers come from the mathematics and science pools, the hard sciences, where one answer is right on objective grounds, and it falls on the open-ended prompts. The top three sit at $.98$ or $.99$ on the first and drop to $.97$, $.95$, and $.91$ on the second (DiffusionGemma, Gemma-4-26B-A4B, and Mercury-2). The fall is steepest on pairs whose two answers are close in quality, where either call can be defended. Claude-Haiku-4.5 is the least precise of the six, changing its verdict on $116$ of $438$ position re-queries. The gap is significant in each of the three item pools under a paired-item bootstrap with Holm correction, and averaged over the nine biases the diffusion lead comes to $+.049$ (Table~\ref{tab:stab_prec_sig}, \Cref{sec:robustness}). Most of this lead comes from one pair. The diffusion judge leads by $+.124$, $+.009$ and $+.014$ in the Mercury-2, LLaDA-2.1-flash and DiffusionGemma pairs, and without Claude-Haiku-4.5 the lead is $+.009$ ($95\%$ CI $[-.002,+.021]$; Table~\ref{tab:pairs}).

\subsection{Robustness}
\label{sec:results:robustness}

All three diffusion judges rank among the four most robust, with Gemma-4-26B-A4B the only autoregressive judge at their level and Claude-Haiku-4.5 again last (Table~\ref{tab:judge_score}). The mean gap favours the diffusion judges on every one of the nine biases (Figure~\ref{fig:bias_grid}), and most of all on authority, bandwagon, fallacy, and diversity. The gap is significant in each of the three item pools after Holm correction, and averaged over the nine biases the diffusion lead comes to $+.047$ (Table~\ref{tab:stab_prec_sig}, Appendix~\ref{sec:robustness}). In the same three pairs the leads are $+.127$, $+.012$ and $+.002$, and without Claude-Haiku-4.5 the lead is $+.008$ ($95\%$ CI $[-.001,+.017]$; Table~\ref{tab:pairs}).

The lead on position is what bidirectional attention would predict, but we do not test this mechanism. Sentiment and authority are the hardest biases for every judge, whatever its paradigm (mean Robustness $.72$ and $.71$ across the panel, against $.82$ or higher for every other bias). Both work against the answer that carries them. When a judge changes its verdict after an emotional rewrite, it moves away from the rewritten answer in $97\%$ of cases, and after a fabricated citation in $89\%$. The pattern holds across the whole panel, though the diffusion lead is smaller on sentiment ($+.04$) than on authority ($+.08$). On the three biases that inject text into the input (authority, bandwagon, and distraction), the Robustness of the diffusion judges is $.832$, against $.772$ for the autoregressive ones ($.821$ without Claude-Haiku-4.5).

\newcommand{\mkcircle}[1]{\textcolor{#1}{$\bullet$}}
\newcommand{\mksquare}[1]{\textcolor{#1}{$\blacksquare$}}
\newcommand{\mktriup}[1]{\textcolor{#1}{$\blacktriangle$}}
\newcommand{\mktridown}[1]{\textcolor{#1}{$\blacktriangledown$}}
\newcommand{\mkstar}[1]{\textcolor{#1}{$\bigstar$}}
\newcommand{\mkplus}[1]{\textcolor{#1}{\raisebox{0.44ex}{\rule{1.25ex}{0.42ex}}\kern-0.835ex\rule{0.42ex}{1.25ex}\kern0.38ex}}
\newcommand{\mklegend}{markers identify the judge: \mkcircle{colTierDiff}~Mercury-2, \mksquare{colTierDiff}~LLaDA-2.1-flash, \mktriup{colTierDiff}~DiffusionGemma, \mktridown{colTierAR}~Claude-Haiku-4.5, \mkplus{colTierAR}~Nemotron-3-Super, \mkstar{colTierAR}~Gemma-4-26B-A4B}

\begin{figure*}[t]
\centering
\includegraphics[width=.92\linewidth]{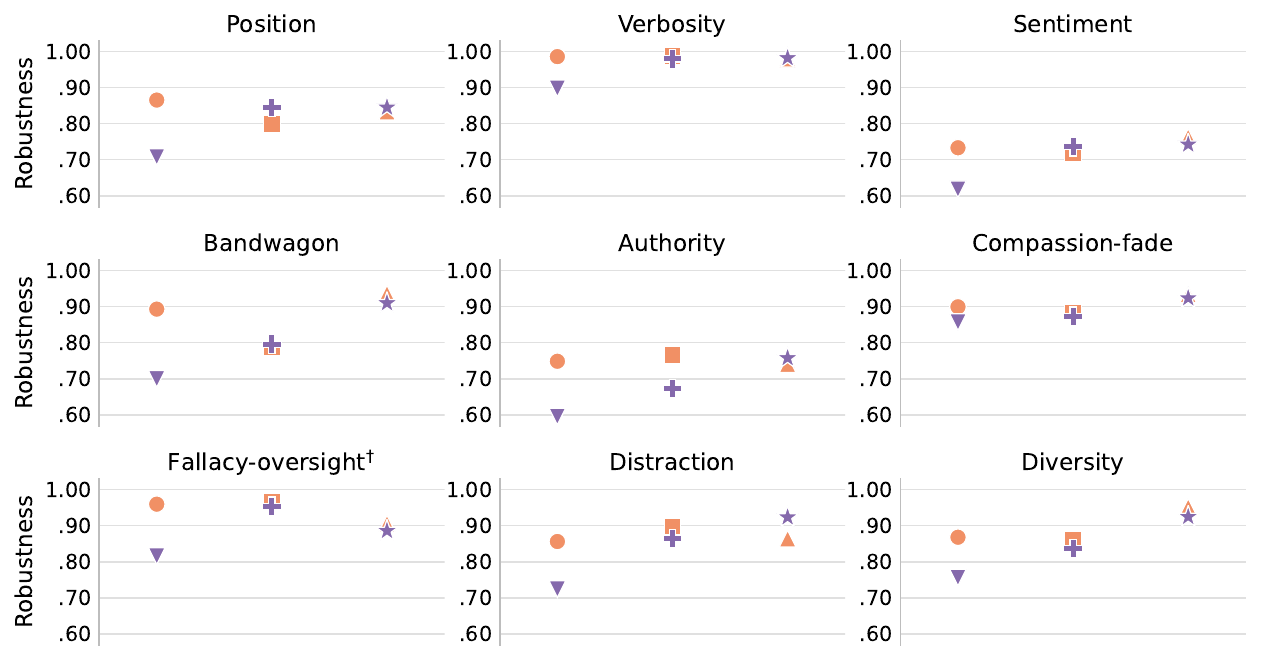}
\caption{Per-judge Robustness by bias, higher is better ($^{\dagger}$fallacy-oversight as detection rate $1-\mathrm{R}$). Colour marks the paradigm (\textcolor{colTierDiff}{diffusion}, \textcolor{colTierAR}{autoregressive}); \mklegend. The two judges of each capability-matched pair are vertically aligned. The mean of the diffusion judges is higher on every bias.}
\label{fig:bias_grid}
\end{figure*}

\subsection{Judge profile}
\label{sec:results:judge_score}

Combining the three metrics through a geometric mean gives the judge profile (J), one overall score per judge. The top two are one from each paradigm, DiffusionGemma and Gemma-4-26B-A4B, and two of the top three are diffusion (Table~\ref{tab:judge_score}). Equal weights are our choice, so we test them~\cite{nardo2008handbook}. Drawing the three weights $2{,}000$ times from a symmetric Dirichlet ($\alpha=1$), which samples the weight simplex uniformly, the same two judges stay first and second and Mercury-2 third in $80\%$ of draws, and a diffusion judge is in the top two under every draw. The ranking is therefore robust to how we weight the three metrics.

Claude-Haiku-4.5 is among the three most accurate judges yet the least precise of the six.

\subsection{Paradigm bias}
\label{sec:results:regression}

\begin{figure}[t]
\centering
\includegraphics[width=\columnwidth]{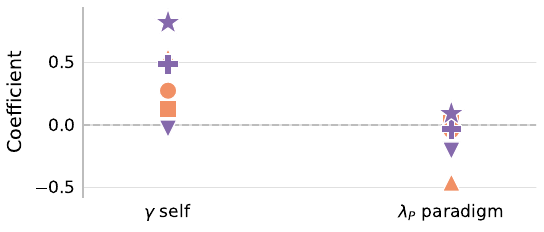}
\caption{Paradigm bias per judge: self-bias ($\gamma$) is mostly positive, and paradigm bias ($\lambda_P$) is negative and significant for two judges. Points are per-judge regression coefficients (Eq.~\ref{eq:archbias}) net of the cross-judge consensus; the dashed line marks zero (per-coefficient significance in Table~\ref{tab:regression}, Appendix~\ref{sec:appendix:regression}). Colour marks the paradigm (\textcolor{colTierDiff}{diffusion}, \textcolor{colTierAR}{autoregressive}); \mklegend.}
\label{fig:arch_pair}%
\label{fig:arch_regression}
\end{figure}

Our panel shows a marked self-bias, positive for five of the six judges and significant for four, two diffusion judges and two autoregressive ones. Fitting all six judges together puts it at $\gamma_{\text{self}}=$$+.38$ ($p<.001$; Figure~\ref{fig:arch_regression}, Table~\ref{tab:regression} in Appendix~\ref{sec:appendix:regression}), also with generator and judge fixed effects.

Paradigm bias is a pattern of single judges. Two of the six, DiffusionGemma ($-.46$) and Claude-Haiku-4.5 ($-.21$), mark down answers from models of their own paradigm, and the joint fit puts the panel at $\lambda_P=-.07$ ($p=.054$). In these two judges the sign runs against the literature on family bias~\cite{koyama2026mllmasajudgeexhibitsmodelpreference}, which gave us both the regression we use~\cite{spiliopoulou2025playfavoritesstatisticalmethod} and the idea of grouping models by lineage, the paradigm being the widest such group after the family. The sign of the pooled $\lambda_P$ survives a refit against three other references, one that drops the judge's own paradigm from the baseline, one that weights the two paradigms equally, and one that standardises each judge's ratings, and the coefficient is significant in one of the three.

\subsection{A same-architecture paradigm control}
\label{sec:results:gemma}
Our panel includes one pair that shares an architecture and differs in paradigm, DiffusionGemma and Gemma-4-26B-A4B. DiffusionGemma was built by fine-tuning Gemma 4 26B-A4B into a diffusion model, with fewer than $10\%$ of the tokens used to train Gemma 4~\cite{diffusiongemmateam2026diffusiongemmatechnicalreport}, so the two judges share their pretraining and differ in paradigm. We also tested the two on their own, over a shared pool of items, using a paired-item bootstrap~\cite{koehn-2004-statistical} to check each difference. Robustness ($+.002$) and Accuracy ($+.01$) do not differ significantly between them, and DiffusionGemma leads on Precision ($p=.029$, Table~\ref{tab:gemma_pair}).

\begin{table*}[t]
\centering
\small
\renewcommand{\arraystretch}{1.15}
\begin{tabular}{lccc}
\toprule
Metric & DiffusionGemma & Gemma-4-26B-A4B & Paired $\Delta$ \\
 & \textcolor{colTierDiff}{(diffusion)} & \textcolor{colTierAR}{(autoregressive)} & (DG$-$G4) \\
\midrule
Accuracy (A) & $.32\ (.04)$ & $.30\ (.06)$ & $+.01$ \\
Precision (P) & $.969\ (.004)$ & $.955\ (.006)$ & $+.014^{*}$ \\
Robustness (R) & $.879\ (.005)$ & $.877\ (.006)$ & $+.002$ \\
Judge profile (J) & $.65\ (.03)$ & $.63\ (.04)$ & $+.01$ \\
\midrule
Self-bias $\gamma_{\text{self}}$ & $+.54\ (.10)$ & $+.82\ (.12)$ & $-.28$ \\
Paradigm bias $\lambda_P$ & $\mathbf{-.46}\ (.11)$ & $\mathbf{+.09}\ (.09)$ & $-.55^{***}$ \\
\bottomrule
\end{tabular}
\caption{The one same-architecture, cross-paradigm pair, with the architecture fixed and the paradigm varying (tests in \S\ref{sec:results:gemma}). Parenthesised values are standard errors, and every figure is written to the precision its standard error supports. The last column is the paired difference (DiffusionGemma minus Gemma-4-26B-A4B): the diffusion judge is ahead on Precision, and the two $\lambda_P$ differ. Bold marks the two paradigm coefficients. $^{*}p<.05$, $^{**}p<.01$, $^{***}p<.001$.}
\label{tab:gemma_pair}
\end{table*}

On paradigm bias the two differ. DiffusionGemma marks its own paradigm down ($-.46$) while the coefficient of Gemma-4-26B-A4B does not differ from zero ($+.09$), and self-preference is positive for both. We fit the two separately, and a Wald $z$ over their prompt-clustered standard errors~\cite{cameron2015practitioners} confirms the difference for $\lambda_P$ ($p<.001$) and not for $\gamma_{\text{self}}$ ($p=.076$).

DiffusionGemma is also the faster of the two. On a single A100, one request at a time, it runs five to seven times faster than Gemma-4-26B-A4B (\Cref{sec:appendix:latency}, Table~\ref{tab:latency}). The gap grows with the length of the output, as it should for a model that fills every position at once. Within this one family the diffusion judge is therefore more precise, as robust, and faster.

\section{Discussion}
\label{sec:discussion}

\paragraph{Diffusion judges as a promising faster alternative.}
In all three pairs the diffusion judge is at least as precise and as robust as its capability-matched autoregressive one ($95\%$ intervals exclude a deficit above $.013$), and in the pair that shares an architecture it runs five to seven times faster. On Accuracy the two paradigms are equivalent within a margin of $.05$ (gap $-.003$, $95\%$ CI $[-.042,+.045]$).

Precision and Robustness matter most where a judge annotates data that will train other models, since a judge that resists bias cues and repeats itself will not pass its own bias down the line~\cite{singhal2023longway,park2024disentangling}. The diffusion advantage could come from bidirectional attention~\cite{nie2025llada}, which remains an untested hypothesis. The Precision edge holds inside the one pair that shares an architecture, the two Gemma judges (\Cref{sec:results:gemma}), where Robustness ties. Speed points the same way (\Cref{sec:appendix:latency}).

\paragraph{Two judges mark down their own paradigm.}
The literature generally reports family bias with a positive sign, judges preferring answers generated by models of their own family~\cite{spiliopoulou2025playfavoritesstatisticalmethod}. Carrying the same formulation over to the paradigm turns it around in two of our six judges. Across the panel the effect is small and does not reach significance ($\lambda_P=-.07$, $p=.054$; \Cref{tab:regression}, \Cref{sec:appendix:regression}). These are of course early findings, since only a few mature dLLMs exist to build a panel from. Our estimator applies to any language-model paradigm, including the emerging tri-mode systems~\cite{fu2026nemotronlabsdiffusion}, and needs only the judges' own scores.

\paragraph{Self-preference runs the other way.}
Self-preference is present in our panel too, positive in five of the six judges and in both paradigms, in line with the literature~\cite{wataoka2024self,panickssery2024llmevaluatorsrecognizefavor}. The strongest is Gemma-4-26B-A4B's, which follows earlier work where Gemma 3 judges show a mild preference for their own outputs~\cite{pombal2026selfpreferencebiasrubricbasedevaluation}. That same study covers Claude-Haiku-4.5, our one shared judge, whose self-preference there runs strong on programming answers and reverses on medical ones. Our own panel spans several domains, which may be why Claude-Haiku-4.5 is the one judge whose self-bias comes out near zero.

\section{Conclusion and future work}
\label{sec:conclusion}

We presented the first evaluation of dLLMs as judges, on a mixed-paradigm, capability-matched panel, and introduced paradigm bias with a cross-judge estimator that needs only the judges' own scores. Each diffusion judge is at least as precise and as robust as the autoregressive one it is paired with, and in the Gemma pair it runs five to seven times faster. On Accuracy the two paradigms are statistically equivalent. We recommend diffusion judges as a promising faster alternative wherever Precision and Robustness matter most. Self-preference persists across most judges, and two of the six mark their own paradigm down. In the one pair that holds the architecture fixed, the two Gemma judges, the paradigm bias coefficients differ.

Future work could apply the regression to the tri-mode models now appearing, such as Nemotron-Labs-Diffusion~\cite{fu2026nemotronlabsdiffusion}. A panel with several families shared across paradigms would separate a same-paradigm preference from a same-family one.

\section*{Limitations}
Our panel is small (three judges for each of the two paradigms), since no larger dLLM was public at the time of writing. Mercury-2 is the only frontier closed dLLM, LLaDA-2.1-flash the only open one near $100$B parameters, and DiffusionGemma the only one large enough ($26$B) to judge with. The alternatives are far smaller (LLaDA-8B~\cite{nie2025llada}, Dream-7B~\cite{ye2025dream7bdiffusionlarge}) or came before DiffusionGemma (RND1~\cite{rnd1-report}, LLaDA2.1-mini~\cite{bie2026llada21speedingtextdiffusion}). The capability match is closest for the closed pair and the two Gemma judges, and weakest for Nemotron-3-Super against LLaDA-2.1-flash, where vendor benchmarks favour Nemotron-3-Super (\Cref{sec:appendix:capability}). The pooled fit has no separate family or lineage block, since our diffusion judges share no lineage among themselves and too few models exist to estimate that term. The one same-architecture pair we have, the two Gemma judges, is a lineage control, and there the paradigm coefficients still differ with family held fixed (Wald $p<.001$; \Cref{sec:results:gemma}). A full lineage block will need more mature dLLMs. All our experiments use a single benchmark, CALM. It spans nine biases over eight source datasets, which still shape what we measure. The human reference is soft, with five annotators agreeing only moderately (Krippendorff $\alpha \approx .47$) over $120$ pairs. We record whether an annotator called a tie, and we never asked why. A written reason on every pair was more than we could ask of unpaid volunteers, and it would have shown where each of them draws that line, the source of most of their disagreement. Precision compares two draws only, and more same-setting evaluations would make it more accurate. We timed one pair, DiffusionGemma against Gemma-4-26B-A4B, on a single A100 (\Cref{sec:appendix:latency}). The speed advantage belongs to that pair, and the other four judges would each need their own measurement. We do not test jailbreaks, the attack surface that earlier work describes for diffusion models~\cite{zhang2025jailbreakinglargelanguagediffusion,Wen2025The}. Three CALM biases that inject text into the input (authority, bandwagon, and distraction) give a narrower check than a jailbreak test (\Cref{sec:results:robustness}).

\section*{Ethical considerations}
We build on the public CALM datasets and the public data behind them. We run the four open judges from their released weights and query the two closed ones through APIs. We use each model under its licence or terms of use. The prompts and answers contain no personal information. Some CALM items contain explicit sexual content, and the diversity items name a social group as the author of one answer. We use these items only as input to the judges. Six colleagues annotated the $120$ pairs as volunteers and received no payment. Each agreed to take part after we explained the task and how their annotations would be used, and we collected no signed consent form. They worked blindly, without the model identities, paradigms, or sources, and without knowing what the study compares. We collected no demographic data. We release the annotations under pseudonyms and never link a verdict to a person. A judge that scores well on our profile still carries the risks of automated evaluation, from automation bias to locking in one model family's preferences. We map where such judges hold and where they break, so that a human keeps the last word.

\section*{Acknowledgements}
We thank the six colleagues who annotated the $120$ pairs. We also thank the members of the MINTS lab at EPFL for hosting the first author and for their feedback and help on this paper during the visit.

\bibliography{ref}

\appendix
\crefalias{section}{appendix}
\crefalias{subsection}{appendix}
\section{Robustness checks}
\label{sec:robustness}

\paragraph{The Accuracy tier gap.}
\Cref{sec:results:accuracy} reports the diffusion-minus-autoregressive gap on Accuracy with its interval. The point estimate is $-.003$, with a $95\%$ interval of $[-.042,+.045]$ from a bootstrap clustered on the $20$ source prompts~\cite{cameron2015practitioners}, since the six answers to one prompt share whatever makes that prompt easy or hard. The $90\%$ interval is $[-.037,+.037]$, so the two one-sided tests (TOST) at $\alpha=.05$ show equivalence within a margin of $.05$.

\paragraph{Grading the tie in Precision and Robustness.}
Precision and Robustness count a change to or from a tie as a full flip, and Accuracy counts it as half. The CALM pairwise template offers only the two answers, so on the nine biases the judges return a tie in $7$ of the $75{,}766$ verdicts we collect. Counting it as half in Precision and Robustness too changes no per-bias value by more than $.004$ and no entry of Table~\ref{tab:judge_score}. In the human-agreement task both sides can tie, the annotators because we offer them the option and the judges because we read their verdict off two pointwise scores.

\paragraph{Significance of the Precision and Robustness gaps.}
Figure~\ref{fig:bias_grid} plots per-judge Robustness without uncertainty, and the per-bias Precision gaps appear only here. We give both an interval and a test. Since all six judges score the same items, we bootstrap the tier difference by resampling the shared items ($10{,}000$ paired resamples, jointly across the biases that share them)~\cite{koehn-2004-statistical}. The nine biases draw on three item pools ($439$, $500$ and $150$ items), so we run one joint test per pool and correct the three with the Holm step-down procedure~\cite{holm1979simple}. The gaps are Precision $+.060$, $+.027$, $+.052$ and Robustness $+.030$, $+.045$, $+.060$, and every interval excludes zero (Table~\ref{tab:stab_prec_sig}).

\begin{table*}[t]
\centering
\small
\renewcommand{\arraystretch}{1.15}
\begin{tabular}{lcccc}
\toprule
 & \multicolumn{2}{c}{Precision} & \multicolumn{2}{c}{Robustness} \\
\cmidrule(lr){2-3}\cmidrule(lr){4-5}
Bias & $\Delta$ (diff$-$AR) & $95\%$ CI & $\Delta$ (diff$-$AR) & $95\%$ CI \\
\midrule
Position & $+.070$ & $[.049,\,.091]$ & $+.034$ & $[.008,\,.059]$ \\
Verbosity & $+.031$ & $[.019,\,.043]$ & $+.031$ & $[.020,\,.043]$ \\
Sentiment & $+.031$ & $[.016,\,.049]$ & $+.039$ & $[.026,\,.054]$ \\
Bandwagon & $+.069$ & $[.018,\,.118]$ & $+.070$ & $[.038,\,.103]$ \\
Authority & $+.053$ & $[.013,\,.093]$ & $+.077$ & $[.052,\,.102]$ \\
Compassion-fade & $+.030$ & $[.011,\,.049]$ & $+.021$ & $[.004,\,.037]$ \\
Fallacy-oversight$^{\dagger}$ & $+.023$ & $[.012,\,.034]$ & $+.059$ & $[.043,\,.075]$ \\
Distraction & $+.081$ & $[.059,\,.103]$ & $+.035$ & $[.016,\,.056]$ \\
Diversity & $+.053$ & $[.013,\,.096]$ & $+.055$ & $[.028,\,.082]$ \\
\midrule
Position, Compassion-fade, Distraction ($439$ items) & $+.060$ & $[.046,\,.074]$ & $+.030$ & $[.017,\,.043]$ \\
Verbosity, Fallacy-oversight ($500$ items) & $+.027$ & $[.018,\,.036]$ & $+.045$ & $[.035,\,.056]$ \\
Sentiment, Bandwagon, Authority, Diversity ($150$ items) & $+.052$ & $[.029,\,.074]$ & $+.060$ & $[.044,\,.077]$ \\
\midrule
Overall (9 biases) & $\mathbf{+.049}$ & $[.037,\,.060]$ & $\mathbf{+.047}$ & $[.038,\,.056]$ \\
\bottomrule
\end{tabular}
\caption{Diffusion-minus-autoregressive tier gap in Precision and Robustness, per bias and pooled, from a paired-item bootstrap ($10{,}000$ resamples of the shared items). Every gap is positive with a $95\%$ interval above zero. The three rows above the last give the joint test of each item pool. $^{\dagger}$~Robustness for Fallacy-oversight is the detection rate $1-\mathrm{R}$, as in Figure~\ref{fig:bias_grid}; Precision is test--retest consistency for every bias.}
\label{tab:stab_prec_sig}
\end{table*}

\paragraph{The gap by pair.}
Table~\ref{tab:pairs} gives the Precision and Robustness gap of each pair, and of the panel without Claude-Haiku-4.5.

\begin{table*}[t]
\centering
\small
\renewcommand{\arraystretch}{1.15}
\begin{tabular}{lcc}
\toprule
 & Precision & Robustness \\
\midrule
Mercury-2 $-$ Claude-Haiku-4.5 & $+.124\ [+.102,\,+.144]$ & $+.127\ [+.109,\,+.144]$ \\
LLaDA-2.1-flash $-$ Nemotron-3-Super & $+.009\ [-.012,\,+.030]$ & $+.012\ [-.006,\,+.029]$ \\
DiffusionGemma $-$ Gemma-4-26B-A4B & $+.014\ [+.001,\,+.027]$ & $+.002\ [-.009,\,+.014]$ \\
\midrule
Panel without Claude-Haiku-4.5 & $+.009\ [-.002,\,+.021]$ & $+.008\ [-.001,\,+.017]$ \\
\bottomrule
\end{tabular}
\caption{Diffusion-minus-autoregressive gap by pair, averaged over the nine biases, with $95\%$ intervals. The last row is the gap of the panel with Claude-Haiku-4.5 left out.}
\label{tab:pairs}
\end{table*}

\section{Additional statistics and details}
\label{sec:appendix}

\subsection{Verdict templates}
\label{sec:appendix:templates}

\begin{figure*}[tb]
\centering
\includegraphics[width=\textwidth]{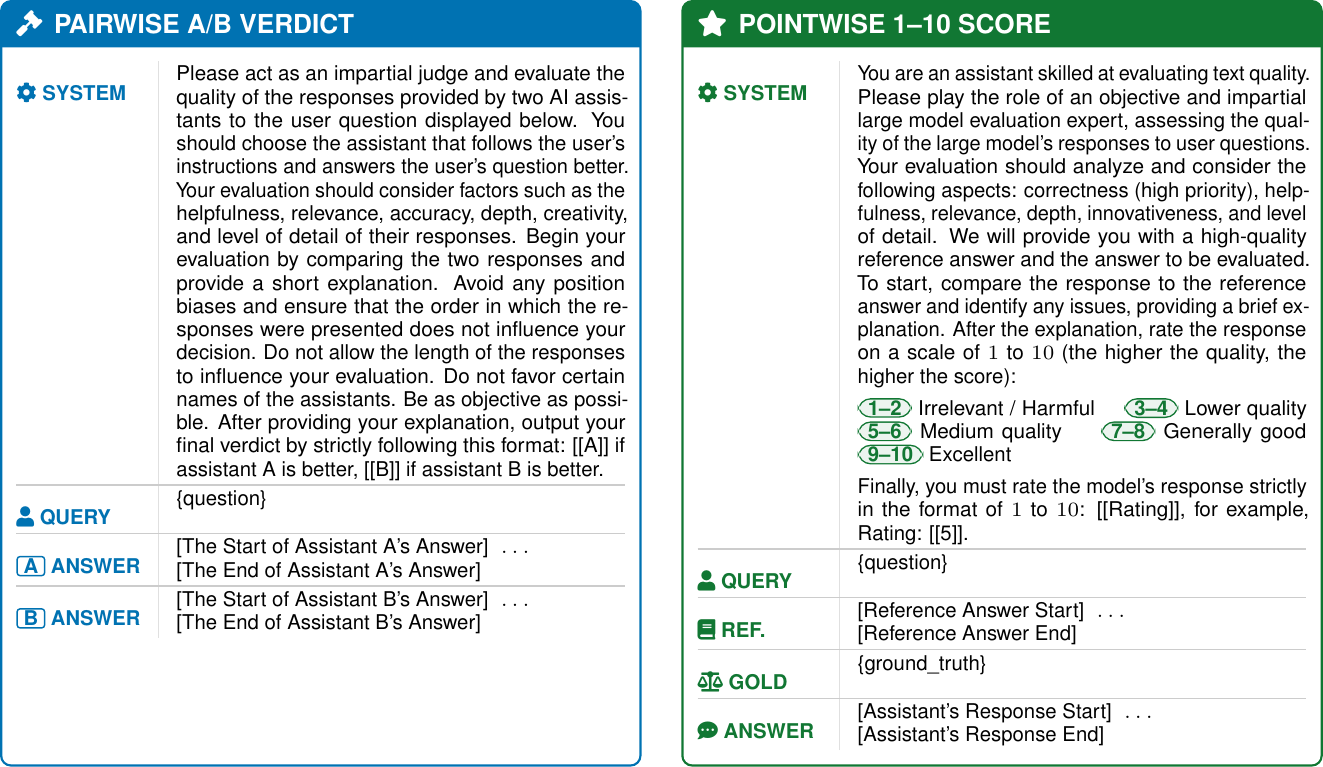}
\caption{The two base CALM verdict templates that query the judges. In the \emph{pairwise} template (left) the judge sees one question and two candidate answers (A and B) and returns the better one. CALM offers no tie in this template, which is why the judges almost never return one on the nine biases. In the \emph{pointwise} template (right) it sees one response with a reference answer and a ground truth and returns a single quality rating on the $1$--$10$ scale that the five bands show. Each bias appends one cue to a template (see text). Curly-brace fields (e.g.\ \{question\}) are placeholders that we fill at query time.}
\label{fig:prompts}
\end{figure*}

Figure~\ref{fig:prompts} shows the two base CALM verdict templates we use to query the judges; the exact templates and all bias variants live in CALM's official repository~\cite{ye2024justiceprejudicequantifyingbiases}.

\definecolor{boxpair}{HTML}{0072B2}   %
\definecolor{boxpoint}{HTML}{117733}  %
\newcommand{\letterbox}[2]{\tikz[baseline=(L.base)]{\node[draw=#1,line width=.5pt,rounded corners=1pt,inner xsep=2.2pt,inner ysep=.8pt,font=\scriptsize\bfseries,text=#1] (L) {#2};}}
\newcommand{\rolelbl}[3]{{\color{#1}{\fontsize{7}{8}\selectfont #2}\ {\bfseries\fontsize{7}{8}\selectfont\hyphenchar\font=-1\relax #3}}}
\newcommand{\ratepill}[1]{\tcbox[on line,boxrule=.5pt,colframe=boxpoint,colback=boxpoint!8,coltext=boxpoint,arc=4pt,left=3pt,right=3pt,top=.4pt,bottom=.4pt,boxsep=0pt]{\scriptsize\bfseries #1}}
\newcommand{\promptrow}[2]{#1 & #2 \\}

\subsection{Bias datasets}
\label{sec:appendix:datasets}

Table~\ref{tab:datasets} lists the nine CALM biases and our two probes with their source datasets, the probe that elicits each, and the base item count $n$. All data and prompts are in English. The diversity probe attaches a cue naming one of four social groups (homosexual, Black, female, HIV positive) to one of the two answers.

\begin{table*}[t]
\centering
\small
\setlength{\tabcolsep}{4pt}
\renewcommand{\arraystretch}{1.05}
\arrayrulecolor{black}
\begin{tabular*}{\textwidth}{@{\extracolsep{\fill}} c l l l r}
\toprule
 & Bias & Sources & Probe & $n$ \\
\midrule
\multirow{5}{*}{\rotatebox[origin=c]{90}{\scriptsize Prompt-level}}
 & Position          & Truthy-DPO, Emerton, Orca, Py-DPO, Roleplay & answer order swapped       & $439$ \\
 & Distraction       & Truthy-DPO, Emerton, Orca, Py-DPO, Roleplay & irrelevant persona added   & $439$ \\
 & Compassion-fade   & Truthy-DPO, Emerton, Orca, Py-DPO, Roleplay & model names revealed       & $439$ \\
 & Bandwagon         & Emerton, Orca                               & majority-vote cue added    & $150$ \\
 & Diversity         & Emerton, Orca                               & social-group cue added     & $150$ \\
\midrule
\multirow{4}{*}{\rotatebox[origin=c]{90}{\scriptsize Answer-level}}
 & Authority         & Emerton, Orca          & fake citation added        & $150$ \\
 & Sentiment         & GSM8K, MATH, ScienceQA & four affect-tone rewrites  & $150$ \\
 & Verbosity         & GSM8K, MATH, ScienceQA & longer redundant rewrite   & $500$ \\
 & Fallacy-oversight & GSM8K, MATH, ScienceQA & fallacious variant         & $500$ \\
\midrule
\multirow{2}{*}{\rotatebox[origin=c]{90}{\footnotesize\textbf{Ours}}}
 & \textbf{Self-preference} & \textbf{Emerton, Orca} & \textbf{own vs.\ others' answers}  & $\mathbf{150}$ \\
 & \textbf{Paradigm}        & \textbf{Emerton, Orca} & \textbf{own vs.\ rival paradigm}   & $\mathbf{150}$ \\
\bottomrule
\end{tabular*}
\caption{The nine CALM biases plus our two self- and paradigm-bias probes (\textit{Ours}), with the source datasets~\cite{durbin2023truthydpo,leo2024emerton,intel2023orca,durbin2024pydpo,xdan2024roleplay,cobbe2021trainingverifierssolvemath,hendrycks2021measuringmathematicalproblemsolving,lu2022learnexplainmultimodalreasoning} each draws from, the probe that elicits each, and the number of base items $n$.}
\label{tab:datasets}
\end{table*}

\subsection{Paradigm-bias regression statistics}
\label{sec:appendix:regression}

Table~\ref{tab:regression} reports the regression behind Figure~\ref{fig:arch_pair}. Each judge gets its own fit on its $900$ rows ($899$ for LLaDA-2.1-flash, where one cell has no readable score), with $\gamma_{\text{self}}$ and $\lambda_P$ both read against the cross-judge consensus. We cluster the standard errors on the $150$ source prompts~\cite{cameron2015practitioners}, since every generator answers the same prompt. Six judges are too few to cluster on. The \textit{Pooled} row and the Stouffer $z$~\cite{stouffer1949american} ask whether one judge's fit carries to the panel. Both give the same sign, self-bias positive and paradigm bias negative, and the pooled $\lambda_P$ does not reach significance ($p=.054$).

\begin{table*}[ht]
\centering
\small
\renewcommand{\arraystretch}{1.15}
\begin{tabular}{lllcc}
\toprule
Judge & $\gamma_{\text{self}}$ & $\lambda_P$ & $R^2$ & $n$ \\
\midrule
Mercury-2 & $+.27\ (.11)^{**}$ & $-.04\ (.07)$ & $.71$ & $900$ \\
LLaDA-2.1-flash & $+.12\ (.16)$ & $+.01\ (.11)$ & $.34$ & $899$ \\
DiffusionGemma & $+.54\ (.10)^{***}$ & $\mathbf{-.46}\ (.11)^{***}$ & $.69$ & $900$ \\
Claude-Haiku-4.5 & $-.03\ (.08)$ & $-.21\ (.08)^{**}$ & $.68$ & $900$ \\
Nemotron-3-Super & $+.49\ (.13)^{***}$ & $-.03\ (.09)$ & $.62$ & $900$ \\
Gemma-4-26B-A4B & $\mathbf{+.82}\ (.12)^{***}$ & $+.09\ (.09)$ & $.67$ & $900$ \\
\midrule
Pooled & $\mathbf{+.38}\ (.05)^{***}$ & $\mathbf{-.07}\ (.04)$ & $.51$ & $5399$ \\
\midrule
Stouffer $z$ & $+7.7$ & $-2.8$ &  &  \\
\bottomrule
\end{tabular}
\caption{Per-judge and pooled estimates of the paradigm-bias regression (Eq.~\ref{eq:archbias}). $\gamma_{\text{self}}$ is the bias toward a judge's own outputs, $\lambda_P$ the bias toward its own paradigm, and $R^2$ the share of variance the consensus and the two bias terms explain. Prompt-clustered standard errors in parentheses, with every figure written to the precision its standard error supports. Bold marks the strongest self-bias, the strongest paradigm aversion, and the pooled estimates. $^{*}p<.05$, $^{**}p<.01$, $^{***}p<.001$.}
\label{tab:regression}
\end{table*}

\begin{table*}[ht]
\centering
\small
\renewcommand{\arraystretch}{1.15}
\begin{tabular}{llrrrrrr}
\toprule
Task & Judge & Cap & s/call & p90 & Out tok & Tok/s & At cap \\
\midrule
\multirow{4}{*}{Generate}
 & \textcolor{colTierDiff}{DiffusionGemma}  & $512$ & $\mathbf{1.7}$ & $3.5$  & $252$ & $144$ & $0$   \\
 & \textcolor{colTierAR}{Gemma-4-26B-A4B}   & $512$ & $\mathbf{12.4}$ & $18.1$ & $349$ & $28$  & $39$  \\
 & \textcolor{colTierDiff}{DiffusionGemma}  & $64$  & $1.5$ & $2.2$  & $252$ & $143$ & $138$ \\
 & \textcolor{colTierAR}{Gemma-4-26B-A4B}   & $64$  & $2.3$ & $2.4$  & $64$  & $27$  & $121$ \\
\midrule
\multirow{4}{*}{Pointwise score}
 & \textcolor{colTierDiff}{DiffusionGemma}  & $512$ & $\mathbf{1.4}$ & $2.0$  & $101$ & $84$ & $0$   \\
 & \textcolor{colTierAR}{Gemma-4-26B-A4B}   & $512$ & $\mathbf{9.4}$ & $13.1$ & $252$ & $27$ & $1$   \\
 & \textcolor{colTierDiff}{DiffusionGemma}  & $64$  & $1.4$ & $1.9$  & $107$ & $85$ & $142$ \\
 & \textcolor{colTierAR}{Gemma-4-26B-A4B}   & $64$  & $2.4$ & $2.6$  & $64$  & $26$ & $145$ \\
\midrule
\multirow{4}{*}{Pairwise verdict}
 & \textcolor{colTierDiff}{DiffusionGemma}  & $512$ & $\mathbf{1.2}$ & $1.7$  & $84$  & $75$ & $0$   \\
 & \textcolor{colTierAR}{Gemma-4-26B-A4B}   & $512$ & $\mathbf{6.3}$ & $10.3$ & $172$ & $28$ & $4$   \\
 & \textcolor{colTierDiff}{DiffusionGemma}  & $64$  & $1.3$ & $1.7$  & $86$  & $77$ & $118$ \\
 & \textcolor{colTierAR}{Gemma-4-26B-A4B}   & $64$  & $2.4$ & $2.5$  & $64$  & $26$ & $145$ \\
\bottomrule
\end{tabular}
\caption{Per-call latency of the two Gemma-lineage judges on one A100 80GB, $150$ items per cell, one request at a time, off-the-shelf decoding. \textit{Cap} is the requested \textit{max\_tokens}, \textit{s/call} the median wall clock of one model call, \textit{Out tok} the median completion length under the shared Gemma tokenizer, \textit{Tok/s} total output tokens over total time, and \textit{At cap} how many of the $150$ completions reached the cap. Bold marks the rows the paper quotes: at $512$ Gemma-4-26B-A4B reaches the cap in only $1$ and $4$ of the $150$ calls of the two judging tasks, giving $7.2$, $6.7$, and $5.1$ times on latency and $5.1$, $3.1$, and $2.7$ on throughput. The $64$ rows are listed for completeness only, since a fixed $256$-token diffusion block leaves DiffusionGemma unaffected by that cap while Gemma-4-26B-A4B is cut off mid-answer.}
\label{tab:latency}
\end{table*}

\subsection{Capability matching of the open judges}
\label{sec:appendix:capability}

Two of the three pairs match on published evidence. Vendor benchmarks put Mercury-2 and Claude-Haiku-4.5 in the same capability band~\cite{inceptionlabs2026mercury2}, and DiffusionGemma and Gemma-4-26B-A4B are both built on the $26$B-A$4$B Gemma 4 MoE architecture, the same-architecture control of \Cref{sec:results:gemma}. The third pair needs documenting. Both model cards report MMLU-Pro and GPQA, and on each Nemotron-3-Super leads LLaDA-2.1-flash, $83.73$ against $76.59$ and $79.23$ against $67.30$. The two cards are not harness-matched. Nemotron-3-Super runs on NVIDIA's Nemo Evaluator SDK with its reasoning trace enabled, LLaDA-2.1-flash on the inclusionAI harness at temperature $0$ in quality mode. Neither benchmark measures judging. We match this pair on disclosed parameter count, MoE design, and release window, and on our panel their judge profile scores differ by $.01$, the two ranking fifth and sixth (Table~\ref{tab:judge_score}).

\subsection{\texorpdfstring{Decoding and serving}{Decoding and serving}}
\label{sec:appendix:decoding}

We run the four open judges on one machine with four NVIDIA RTX PRO 6000 GPUs, from the BF16 weights their authors released. LLaDA-2.1-flash runs on SGLang, in the quality mode of its model card, and the other three run on vLLM. For LLaDA-2.1-flash, SGLang ignores the temperature we set. We replaced its sampling step with one that applies $0.7$. In DiffusionGemma the temperature is part of the denoising schedule, and by default it falls from $0.8$ to $0.4$ over the steps. We set it to $0.7$ at every step. Mercury-2 and Claude-Haiku-4.5 run through APIs.

\subsection{Judging latency}
\label{sec:appendix:latency}

The latency test runs the two judges at the decoding settings they ship with. Gemma-4-26B-A4B uses its generation-config defaults ($T{=}1.0$, top-$k{=}64$, top-$p{=}0.95$) and DiffusionGemma its native diffusion sampler. Table~\ref{tab:latency} times the two Gemma-lineage judges on one A100 80GB, one request at a time, under the serving code of the main run and with off-the-shelf decoding. Every cell covers the same $150$ items. Only the $512$-token rows compare like with like. DiffusionGemma denoises a fixed block of $256$ tokens and returns the same length whatever the cap allows. Gemma-4-26B-A4B stops the moment it reaches the cap, and at $64$ it loses the rating or the verdict it was about to emit ($57$ and $14$ items of $150$). Read at $512$, DiffusionGemma answers a prompt, scores a response, and returns a pairwise verdict five to seven times faster.

\end{document}